\documentclass{article}

\usepackage{arxiv}

\usepackage[utf8]{inputenc} 
\usepackage[T1]{fontenc}    
\usepackage{hyperref}       
\usepackage{url}            
\usepackage{booktabs}       
\usepackage{amsfonts}       
\usepackage{nicefrac}       
\usepackage{microtype}      
\usepackage{lipsum}
\usepackage{amsmath}
\usepackage{algorithm}
\usepackage{algorithmicx}
\usepackage{algpseudocode}
\usepackage{graphicx}  
\usepackage{lineno}
\usepackage{multirow}
\title{Learning to Transfer: \\Unsupervised Meta Domain Translation}

\author{Jianxin Lin\textsuperscript{1}~~~~~~~~Yijun Wang\textsuperscript{1}~~~~~~~~Tianyu He\textsuperscript{1}~~~~~~~~Zhibo Chen\textsuperscript{1}\\
            \textsuperscript{1}University of Science and Technology of China\\
            \texttt{\{linjx,wyjun,hetianyu\}@mail.ustc.edu.cn}~~~~~~~~\texttt{chenzhibo@ustc.edu.cn}\\
            }

\begin{document}
\maketitle

\begin{abstract}
Unsupervised domain translation has recently achieved impressive performance with Generative Adversarial Network (GAN) and sufficient (unpaired) training data. However, existing domain translation frameworks form in a disposable way where the learning experiences are ignored and the obtained model cannot be adapted to a new coming domain. In this work, we take on unsupervised domain translation problems from a meta-learning perspective. We propose a model called Meta-Translation GAN (MT-GAN) to find good initialization of translation models. In the meta-training procedure, MT-GAN is explicitly trained with a primary translation task and a synthesized dual translation task. A cycle-consistency meta-optimization objective is designed to ensure the generalization ability. We demonstrate effectiveness of our model on ten diverse two-domain translation tasks and multiple face identity translation tasks. We show that our proposed approach significantly outperforms the existing domain translation methods when each domain contains no more than ten training samples.
\end{abstract}

\section{Introduction}
Unsupervised domain translation tasks~\cite{zhu2017unpaired,choi2017stargan}, which aim at learning a mapping that can transfer images from a source domain to a target domain using unpaired training data only, have been widely investigated in recent years. However, current literature focuses on learning a model for a specific translation task without considering the generalization ability to other tasks. In comparison, human intelligence has the ability to quickly learn new concepts with the prior experiences. Taking painting as an example, after being taught how to paint in Monet's style, people have learned the basic skills of painting with usage of painting brush, palette, etc. When being required to learn painting in Van Gogh's style, we do not need to learn from scratch for all these skills. Instead, we can quickly adapt to this new task by viewing only a few Van Gogh's paintings, extracting his painting style, combining the basic painting knowledge we learned before with the new style and eventually know how to draw in Van Gogh's way.

In this paper, we take a step towards unsupervised domain translation (UDT) problems from a meta-learning perspective, aiming to effectively leverage learning experiences from previous domain translation tasks. Briefly, the problem is that we have several domain translation tasks where each tasks only consists of limited samples. We have to learn a meta-model from these tasks, that can be quickly adapted to unseen domain translation tasks. To tackle this problem, we propose a method called Meta-Translation Generative Adversarial Network (MT-GAN) that are robust to different task contexts.
The proposed model contains two meta-learners: a meta-generator $G$ which keeps the memory of prior translation experiences and a meta-discriminator $D$ which teaches $G$ how to quickly generalize to a new task.  Our approach leverages both model-agnostic meta-learning algorithm (MAML) \cite{finn2017model} and Generative Adversarial Network (GAN) \cite{goodfellow2014generative} to iteratively update $G$ and $D$. To achieve that, within a meta-training iteration, for a specific translation task, we synthesize its dual translation task with current states of MT-GAN and train these two tasks in a dual learning form \cite{zhu2017unpaired,he2016dual}. Then we design a meta-optimization objective to evaluate the performance of fine-tuned MT-GAN, and minimize the expected losses on the meta-testing samples with respect to parameters of MT-GAN, which ensures that the direction taken to fine-tuning leads to a good generalization performance.


We extensively evaluate the effectiveness and generalizing ability of the proposed MT-GAN algorithm on two kinds of translation task distributions. The first one contains $10$ diverse two-domain translation tasks covering a wide range of scenarios, including labels$\leftrightarrow $photos, horses$\leftrightarrow $zebras, summer$\leftrightarrow $winter, etc.  The second one is established by repeatedly sampling two arbitrary identities, which forms a two-domain translation task, from a multiple identity dataset \cite{ng2014data}. For a translation task $T$, we take the other $9$ different tasks as the training dataset and test the meta-learned parameter initialization on task $T$. Our experiments only use $10$ samples at most in an image domain of a translation task and show that the proposed meta-learning approach outperforms ordinary domain translation models, such as CycleGAN \cite{zhu2017unpaired} and StarGAN \cite{choi2017stargan}.

We summarize our contributions as follow: 1) We approach the unsupervised domain translation problem from a meta-learning perspective, which allows effective usage of learning experiences from previous domain translation tasks when taking up a new domain translation task; 2) We propose a method (MT-GAN) that jointly trains two meta-learners in an adversarial and dual form, which to the best of our knowledge has not been explored; 3) We extensively verify the effectiveness of our meta-learning based approach on a wide range of translation tasks.

\section{Related works}
\textbf{Generative Adversarial Network}
In recent years, the generative adversarial network (GAN) model \cite{goodfellow2014generative} has gained a wide range of interests in generative modeling.  In a GAN, a generator is trained to produce fake but plausible images, while a discriminator is trained to distinguish difference between real and fake images. The conditional version of a GAN, called a conditional generative adversarial network (CGAN) \cite{mirza2014conditional}, is a model in which the generator is feeded with noise vector together with additional data (e.g., class labels) that conditions on both the generator and discriminator. Deep convolutional generative adversarial network (DCGAN)  \cite{radford2015unsupervised} is an extensive exploration of convolution neural network architectures in GANs and contributes to improve the quality of image synthesis. GANs have been successfully leveraged in many image generation applications \cite{pumarola2018ganimation,wang2018high,wu2016learning,yu2018generative}. Our method adopts the adversarial loss to render images from the generators to be close to real in the target domain and make meta-training performance improve meta-learners' generalization.

\textbf{Unsupervised Domain Translation}
Domain translation has also achieved impressive performance thanks to recent development of GANs and availability of sufficient training data. Isola et al. \cite{isola2016image} proposed a general conditional GAN (Pix2Pix) framework for a wide range of supervised domain translation tasks. Since obtaining an amount of paired training data can be difficult and impractical for many domain translation tasks, DualGAN~\cite{Yi_2017_ICCV}, DiscoGAN~\cite{kim2017learning} and CycleGAN~\cite{zhu2017unpaired} were proposed to learn two cross-domain translation models that obey the cycle consistent rule from unpaired data.  Choi et al. \cite{choi2017stargan} further proposed a unified unpaired domain translation model (StarGAN) to perform domain translation for multiple domains. Liu et al. \cite{liu2018unified} also proposed a method called UFDN to learn domain-invariant representation for multiple domain translation and can perform diverse domain translation and manipulation. A related work  similar to our work may be Benaim et al. \cite{benaim2018one}, in which they proposed a one-shot cross-domain translation which transfers one and only one image in a source domain to a target domain with sufficient data in the target domain. However, all these existing unsupervised domain translation models mainly rely on training data of current translation task, and omit to utilize the meta-knowledge from prior learning experiences like humans. In this work, we focus on image translation that incorporates the prior learning experiences from other translation tasks  for new translation tasks' learning.

\textbf{Meta-Learning}
Meta-learning, which aims to learn a particular process to adjust meta-learners that perform well on a new task, can be traced back to early works \cite{schmidhuber1987evolutionary,bengio1990learning,bengio1992optimization}. Some recent meta-learning studies have focused on learning a shared metric by comparing similarity among data samples. Specifically, Vinyals et al. \cite{vinyals2016matching} proposed Matching Networks that learns an embedding function and measures similarity using the cosine distance in an attention kernel. Snell et al. \cite{snell2017prototypical} also proposed to compare new examples in a learned metric space but used the Euclidean distance with a linear classifier. Another popular approach to meta-learning is to learn a shared initialization of network parameters. For example, Finn et al. \cite{finn2017model} presented model-agnostic meta-learning (MAML) to optimize the parameters of a meta-learner with the objective of maximizing its performance on a new task after a small number of gradient steps. Several other methods \cite{andrychowicz2016learning,mishra2017simple} utilized an additional memory-based network (e.g., LSTM) as the meta-learner. Observing that meta-learning methods usually require labeled datasets, recent works \cite{hsu2018unsupervised,metz2018meta} also proposed to tackle the unsupervised meta-learning. In this paper, we extend the concept of meta-learning to image translation. Specifically, we jointly train two meta-learners in an adversarial and dual form, which to the best of our knowledge has not been done before.


\section{Unsupervised Domain  Translation via Meta-Learning}

\subsection{Problem Formulation}
Domain Translation is the problem of finding a meaningful correspondence between two domains. Since paired supervision is not available in a majority of settings, many works focus on Unsupervised Domain Translation (UDT) where data samples from each domain are unpaired \cite{de2019optimal}.
Existing unsupervised domain translation models mainly rely on training data of a specific translation task, lacking the ability of utilizing the knowledge from prior learning experiences when taking up alternative translation tasks.
To tackle this problem, we approach the unsupervised domain translation problem from a meta-learning perspective, which allows effective usage of learning  experiences from previous domain translation tasks.
The goal of our method is to first leverage unpaired data for efficient training, and then the obtained model can be applied on a wide range of new domain translation tasks.

Formally, in our meta-learning scenario, assuming that there are a series of tasks following distribution $P(T)$ over the task space $\mathcal{T}$. For a specific task $T \sim P(T)$, intuitively task $T$ aims at finding a meaningful correspondence between two domains. Specifically, both domains can be expressed as probability distributions $P_T(x)$ and $P_T(y)$ supported over domain spaces $\mathcal{X}_T$ and $\mathcal{Y}_T$. A task consists in finding a mapping: $\mathcal{X}_T \rightarrow \mathcal{Y}_T$ such that the mapping yields semantically meaningful pairings.

Since we aims at utilizing the learning experience from previous domain translation tasks when take up new translation tasks, in our meta-learning scenario, the training sample consists of a finite set of tasks $\{T_n\}^N_{n=1}$ drawn from $P(T)$, where $N$ denotes the number of training tasks. Concretely, each training task $T_n$ is a tuple $T_n = (S_{T_n} ,Q_{T_n})$, where  $S_{T_n}$ denotes the support set and $Q_{T_n}$ denotes the query set. Specifically, the support set $S_{T_n}= \{\{x_k\}^K_{k=1}\in \mathcal{X}_{T_n}, \{y_k\}^K_{k=1}\in \mathcal{Y}_{T_n}\}$ contains $2K$ unpaired samples from two different domains $\mathcal{X}_{T_n}$ and $\mathcal{Y}_{T_n}$. $Q_{T_n}= \{\{x_l\}^L_{l=1}\in \mathcal{X}_{T_n}, \{y_l\}^L_{l=1}\in \mathcal{Y}_{T_n}\}$ is the query set that contains $2L$ unpaired samples from the same two domains. $S_{T_n}$ and $Q_{T_n}$ are disjoint. Our algorithm takes $\{T_n\}^N_{n=1}$ as inputs and produces a learning strategy for two meta-learners, i.e., $G$ and $D$. In general, the meta-learners iteratively adjust the parameters on data from support set and assess their generalization performance by calculating meta-objective with data from query set. The meta-learners are then improved by considering how the test error  changes with respect to the parameters. In effect, the test error on sampled tasks $T_n$ serves as the training error of the meta-learning process.
At inference time, suppose that we have a new translation task $T_{N+1}=(S_{T_{N+1}} ,Q_{T_{N+1}})$, where $S_{T_{N+1}}=\{\{x_k\}^K_{k=1}\in \mathcal{X}_{T_{N+1}}, \{y_k\}^K_{k=1}\in \mathcal{Y}_{T_{N+1}}   \}$ and $Q_{T_{N+1}}=\{\{x_l\}^L_{l=1}\in \mathcal{X}_{T_{N+1}}, \{y_l\}^L_{l=1}\in \mathcal{Y}_{T_{N+1}}   \}$ ($S_{T_{N+1}}$ and $Q_{T_{N+1}}$ are disjoint.), the learning strategy should learn a fine-tuned $G$ and a fine-tuned $D$ with $S_{T_{N+1}}$ and accomplishes the translation $X \rightarrow Y$ on $Q_{T_{N+1}}$. That is, the meta-performance is measured by the meta-learners' performance after learning from $K$ samples in each domain. We denote the above process as $K$-shot domain translation.

\subsection{Our Approach}

We introduce the formulation of MT-GAN as following: for a $K$-shot domain translation problem and dataset $\{T_n\}^N_{n=1}$, our goal is to find a meta-generator $G$ that keeps the memory of prior translation experiences and a meta-discriminator $D$ that teaches $G$ how to quickly generalize to a new task. Unlike CycleGAN or StarGAN that utilizes multiple generators (and discriminators) or multiple domain codes to tackle different translation tasks, we only develop one meta-generator $G$ and one meta-discriminator $D$ for different tasks. This is because a initialization is responsible to the learning strategy of all translation tasks and is independent with a specific task.
The full algorithm of MT-GAN is outlined in Algorithm \ref{alg:MT-GAN} in a general case.

Formally, the generator $G$ and discriminator $D$ are parameterized by $\theta_g$ and $\theta_{d}$ respectively. In each meta-batch, we sample $J$ tasks for training. For a specific translation task $T \sim P(T)$ in one meta-batch, $T = (S_{T} ,Q_{T})$, there are two image domains $\mathcal{X}_T$ and $\mathcal{Y}_T$. Our primary target is to learn a mapping $F$: $\mathcal{X}_T \rightarrow \mathcal{Y}_T$ that is derived from $G$. Since there is no paired data for training, observing that there is naturally a dual task which learns another mapping $H$: $\mathcal{Y}_T\rightarrow \mathcal{X}_T$ in the reverse direction, we utilize these two translation tasks as a two-agent game for fine-tuning in the meta-training period. Specifically, two discriminators $D_\mathcal{Y}$ and $D_\mathcal{X}$ are used to render images from the generators to be real in corresponding domain.

\begin{algorithm}[t]
	\caption{MT-GAN training process}
	\label{alg:MT-GAN}
	\begin{algorithmic}[1]
		\Require: Distribution $P(T)$ over domain translation tasks		
		\Require: Hyperparameters $\alpha$, $\beta$, $\lambda_{\text{cyc}}$, $\lambda_{\text{idt}}$, $K$, $J$
		\State Randomly initialize parameters $\theta_{g}$ of $G$ and $\theta_{d}$ of $D$
		\State \textbf{while} not converged \textbf{do}
		\State $\quad$ Sample batch of tasks $\{T_j\}_{j=1}^J \sim P(T)$, where $J$ is the meta-batch size.
		\State $\quad$ Split support set and query set:   $S_{T_j}$ and $Q_{T_j} \leftarrow T_j$
		\State $\quad$ \textbf{for all} $S_{T_j}$ \textbf{do}
		\State $\quad$ $\quad$ \textbf{Meta-training}:
		\State $\quad$ $\quad$ Compute initialized parameters ${\theta'}_{d_{\mathcal{X}},0}$, ${\theta'}_{d_{\mathcal{Y}},0}$,\\ $\quad$ $\quad$ ${\theta'}_{f,0}$, ${\theta'}_{h,0}$ with
		 gradient descent by Eqn.(\ref{eq:update_D0}) \\ $\quad$ $\quad$ and Eqn.(\ref{eq:update_G0})
		\State $\quad$ $\quad$ \textbf{for} $i$ \textbf{in} iterations $I$ \textbf{do}
		\State $\quad$ $\quad$ $\quad$ Compute fine-tuned parameters ${\theta'}_{d_\mathcal{X},i+1}$,\\ $\quad$ $\quad$ $\quad$ ${\theta'}_{d_\mathcal{Y},i+1}$, ${\theta'}_{f,i+1}$, ${\theta'}_{h,i+1}$ with gradient descent \\ $\quad$ $\quad$ $\quad$ by Eqn.(\ref{eq:update_Dt}) and Eqn.(\ref{eq:update_Gt})
		\State $\quad$ $\quad$ \textbf{end for}
		\State $\quad$ $\quad$ \textbf{Meta-testing}:
		\State $\quad$ $\quad$ Compute meta-objective $\mathcal{L}_{T_j,I+1}^q$ on  $Q_{T_j}$ \\ $\quad$ $\quad$ according to Eqn.(\ref{eq:meta_loss})
		\State $\quad$ \textbf{end for}
		\State $\quad$ \textbf{Meta-Optimization}:
		\State $\quad$ Update ${\theta}_{d}={\theta}_{d}+\beta\nabla_{{\theta}_{d}}\sum_{T_j \in \{T_j\}_{j=1}^J} \mathcal{L}_{T_j,I+1}^q$
		\State $\quad$ Update ${\theta}_{g}={\theta}_{g}-\beta\nabla_{{\theta}_{g}}\sum_{T_j \in \{T_j\}_{n=1}^J} \mathcal{L}_{T_j,I+1}^q$
		\State \textbf{end while}
	\end{algorithmic}
\end{algorithm}

In the initialization step, we initialize the parameters of discriminators and generators as:
\begin{equation}
	\begin{aligned}
		{\theta'}_{d_{\mathcal{Y}},0}&=\theta_d+\alpha\nabla_{\theta_d}\mathcal{L}_{T,0}; \\{\theta'}_{d_{\mathcal{X}},0}&=\theta_{d,wc}+\alpha\nabla_{\theta_{d,wc}}\mathcal{L}_{T,0},
		\label{eq:update_D0}
	\end{aligned}
\end{equation}
\begin{equation}
	\begin{aligned}
		{\theta'}_{f,0}&=\theta_g-\alpha\nabla_{\theta_g}\mathcal{L}_{T,0}; \\{\theta'}_{h,0}&= \theta_{g,wc}-\alpha\nabla_{\theta_{g,wc}}\mathcal{L}_{T,0},
		\label{eq:update_G0}
	\end{aligned}
\end{equation}
where $\alpha$ is the learning rate during the meta-training period; ${\theta'}_{d_{\mathcal{Y}},0}$, ${\theta'}_{f,0}$, ${\theta'}_{d_{\mathcal{X}},0}$ and ${\theta'}_{h,0}$ are the parameters of $D_{\mathcal{Y}}$, $F$, $D_{\mathcal{X}}$ and $H$ respectively after initialization; $\theta_{d,wc}$ and $\theta_{g,wc}$ are the parameters of $wc(D)$ and $wc(G)$, where $wc$ is the network weights copy operation that detaches back-propagation gradient from meta-optimization objectives. $G$ and $D$ can be parameter initialization for any translation task in practice. However, in a specific task $T$, using $G$ for fine-tuning of both $F$ and $H$ is ambiguous since the meta-optimization objective would become to require $G$ and $D$ to be well adapted for both $\mathcal{X}_T\rightarrow \mathcal{Y}_T$ and $\mathcal{Y}_T\rightarrow \mathcal{X}_T$ at the same time. Therefore, we update $G$ and $D$ only for $\mathcal{X}_T\rightarrow \mathcal{Y}_T$ translation with $wc$ operation, and utilize $\mathcal{Y}_T \rightarrow \mathcal{X}_T$ translation as a dual task. The overall objective $\mathcal{L}_{T,0}$ for training the discriminators and generators at initialization step is given as:
\begin{equation}
	\begin{aligned}
		\mathcal{L}_{T,0}(G, D, \mathcal{X}_T, \mathcal{Y}_T) = & \mathcal{L}_{\text{adv}}(G, D, \mathcal{X}_T, \mathcal{Y}_T) \\+& \mathcal{L}_{\text{adv}}(wc(G), wc(D), \mathcal{Y}_T, \mathcal{X}_T ) \\ +& \lambda_{\text{cyc}}\mathcal{L}_{\text{cyc}}(G, wc(G), \mathcal{X}_T, \mathcal{Y}_T) \\ +& \lambda_{\text{idt}}\mathcal{L}_{\text{idt}}(G, wc(G), \mathcal{X}_T, \mathcal{Y}_T),
		\label{eq:total_loss0}
	\end{aligned}
\end{equation}
where the factors $\lambda_{\text{cyc}}$ and $\lambda_{\text{idt}}$ are used to balance different loss terms $\mathcal{L}_{\text{adv}}$, $\mathcal{L}_{\text{cyc}}$ and $\mathcal{L}_{\text{idt}}$. Specifically, the loss terms $\mathcal{L}_{\text{adv}}$, $\mathcal{L}_{\text{cyc}}$ and $\mathcal{L}_{\text{idt}}$  are the adversarial loss, the cycle-consistency loss and the identity loss respectively, which are defined as follows:
\begin{equation}
	\begin{aligned}
		\mathcal{L}_{\text{adv}}(G, D, \mathcal{X}_T, \mathcal{Y}_T) =& \mathbb{E}_{y\sim P_T(y)}[\log D(y)] \\ +&\mathbb{E}_{x\sim P_T(x)}[\log(1-D(G(x)))],
		\label{eq:adv_loss}
	\end{aligned}
\end{equation}
\begin{equation}
	\begin{aligned}
		\mathcal{L}_{\text{cyc}}(G_1, G_2, \mathcal{X}_T, \mathcal{Y}_T) =& \mathbb{E}_{x\sim P_T(x)}[\Vert G_2(G_1(x))-x\Vert_1] \\ +& \mathbb{E}_{y\sim P_T(y)}[\Vert G_1(G_2(y))-y\Vert_1],
		\label{eq:cyc_loss}
	\end{aligned}
\end{equation}
\begin{equation}
	\begin{aligned}
	\mathcal{L}_{\text{idt}}(G_1, G_2, \mathcal{X}_T, \mathcal{Y}_T) =& \mathbb{E}_{x\sim P_T(x)}[\Vert G_2(x)-x\Vert_1] \\ +& \mathbb{E}_{y\sim P_T(y)}[\Vert G_1(y)-y\Vert_1],
	\label{eq:idt_loss}
	\end{aligned}
\end{equation}
where $P_T(x)$ and $P_T(y)$ are the distributions of samples in domain spaces $\mathcal{X}_T$ and $\mathcal{Y}_T$ respectively. Then, we estimate the expectation terms in above equations using the samples in support set $S_T$.

At iteration $i\geq 0$, we follow the popular unsupervised domain translation model, e.g., CycleGAN, to fine-tune the initialized $D_\mathcal{X}$, $D_\mathcal{Y}$, $F$ and $H$ to quickly adapt to the task $T$. Formally, we update the parameters as follow:

\begin{equation}
	\begin{aligned}
	&{\theta'}_{d_\mathcal{Y},i+1}={\theta'}_{d_\mathcal{Y},i}+\alpha\nabla_{{\theta'}_{d_\mathcal{Y},i}}\mathcal{L}_{T,i+1};\\ \quad &{\theta'}_{d_\mathcal{X},i+1}={\theta'}_{d_\mathcal{X},i}+\alpha\nabla_{{\theta'}_{d_\mathcal{X},i}}\mathcal{L}_{T,i+1},
	\label{eq:update_Dt}
	\end{aligned}
\end{equation}
\begin{equation}
	\begin{aligned}
	&{\theta'}_{f,i+1}={\theta'}_{f,i}-\alpha\nabla_{{\theta'}_{f,i}}\mathcal{L}_{T,i+1};\\ \quad &{\theta'}_{h,i+1}={\theta'}_{h,i}-\alpha\nabla_{{\theta'}_{h,i}}\mathcal{L}_{T,i+1}.
	\label{eq:update_Gt}
	\end{aligned}
\end{equation}
The training objective $\mathcal{L}_{T,i+1}$ to fine-tune the generators and discriminators is given as:

\begin{equation}
	\begin{aligned}
		&\mathcal{L}_{T,i+1}(F_i, H_i, D_{\mathcal{X},i}, D_{\mathcal{Y},i}, \mathcal{X}_T, \mathcal{Y}_T)\\
		=& \mathcal{L}_{\text{adv}}(F_i, D_{\mathcal{Y},i},  \mathcal{X}_T, \mathcal{Y}_T) + \mathcal{L}_{\text{adv}}(H_i, D_{\mathcal{X},i}, \mathcal{Y}_T, \mathcal{X}_T) \\
		+& \lambda_{\text{cyc}}\mathcal{L}_{\text{cyc}}(F_i, H_i, \mathcal{X}_T, \mathcal{Y}_T) + \lambda_{\text{idt}}\mathcal{L}_{\text{idt}}(F_i, H_i, \mathcal{X}_T, \mathcal{Y}_T),
		\label{eq:total_lossT}
	\end{aligned}
\end{equation}
where $D_{\mathcal{Y},i}$, $F_i$, $D_{\mathcal{X},i}$ and $H_i$ are the state of the $D_{\mathcal{Y}}$, $F$, $D_{\mathcal{X}}$ and $H$ at meta-training iteration $i$. We estimate the expectation terms in $\mathcal{L}_{T,i+1}$ also using the samples in support set $S_T$.

For meta-optimization, we minimize the expected loss on query set $Q_T$ with updated discriminators and generators across the task $T$ to train the initial parameters of $D$ and $G$. Our MT-GAN model can be trained as follows:

\begin{align}
	&{\theta}_{d}={\theta}_{d}+\beta\nabla_{{\theta}_{d}}\mathcal{L}^q_{T,I+1},
	\label{eq:update_DT}\\
	&{\theta}_{g}={\theta}_{g}-\beta\nabla_{{\theta}_{g}}\mathcal{L}^q_{T,I+1},
	\label{eq:update_GT}
\end{align}
where $\beta$ is the learning rate for the meta-optimization, and $I$ is the overall iteration number of meta-training. The overall meta-optimization objective $\mathcal{L}^q_{T,I+1}$ for the meta-generator and meta-discriminator is given as:
\begin{equation}
	\begin{aligned}
		&\mathcal{L}^q_{T,I+1}(F_I, H_I, D_{\mathcal{X},I}, D_{\mathcal{Y},I}, \mathcal{X}_T,\mathcal{Y}_T)\\
		=& \mathcal{L}_{\text{adv}}(F_I, D_{\mathcal{Y},I}, \mathcal{X}_T, \mathcal{Y}_T) + \mathcal{L}_{\text{adv}}(H_I, D_{\mathcal{X},I}, \mathcal{Y}_T, \mathcal{X}_T) \\
		+& \lambda_{\text{cyc}}\mathcal{L}_{\text{cyc}}(F_I, H_I, \mathcal{X}_T, \mathcal{Y}_T) + \lambda_{\text{idt}}\mathcal{L}_{\text{idt}}(F_I, H_I, \mathcal{X}_T, \mathcal{Y}_T),
		\label{eq:meta_loss}
	\end{aligned}
\end{equation}
where we estimate the expectation terms in $\mathcal{L}^q_{T,I+1}$ using the samples in query set $Q_T$.

At inference time, for an unseen translation task $T_{N+1}=(S_{T_{N+1}} ,Q_{T_{N+1}})$, where $S_{T_{N+1}}=\{\{x_k\}^K_{k=1}\in \mathcal{X}_{T_{N+1}}, \{y_k\}^K_{k=1}\in \mathcal{Y}_{T_{N+1}}   \}$ and $Q_{T_{N+1}}=\{\{x_l\}^L_{l=1}\in \mathcal{X}_{T_{N+1}}, \{y_l\}^L_{l=1}\in \mathcal{Y}_{T_{N+1}}   \}$, we iteratively fine-tune the obtained $G$ and $D$ with meta-training steps from Eqn.(\ref{eq:update_D0}) to Eqn.(\ref{eq:total_lossT}) to obtain an $F$ that transfers $\{x_k\}^L_{l=1}$ to $\mathcal{Y}_{T_{N+1}}$ domain and an $H$ that transfers $\{y_k\}^L_{l=1}$ to $\mathcal{X}_{T_{N+1}}$ domain.

\begin{figure*}[htb!]
	\centering
	\centerline{\includegraphics[width=13.5cm]{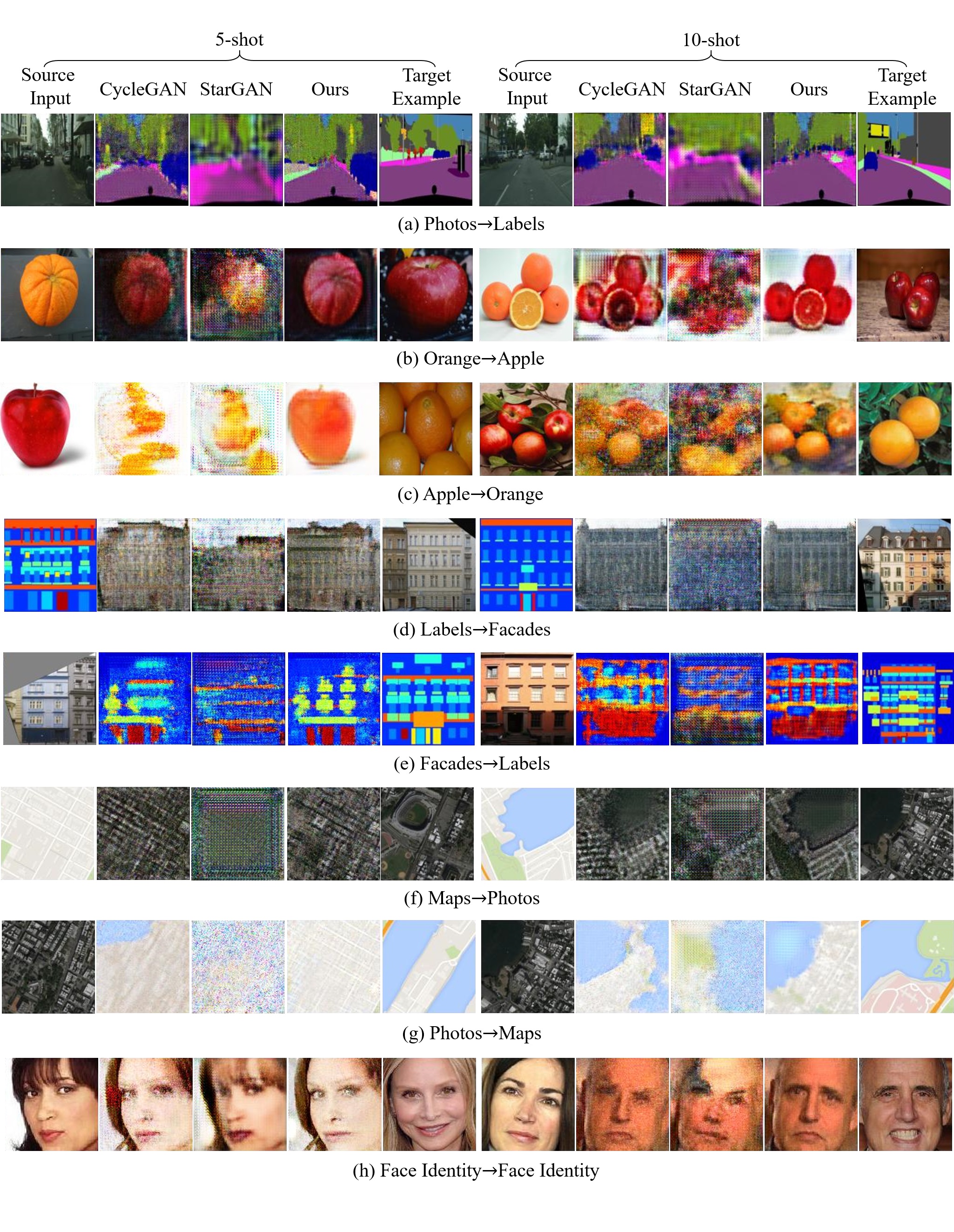}}
	\caption{ Translation results of various testing tasks using CycleGAN, StarGAN and our MT-GAN. Left part: 5-shot domain translation. Right part: 10-shot domain translation. From left to right, the columns represent inputs from the source domain, CycleGAN's results, StarGAN's results, our results and examples in the target domain respectively.}
	\label{fig:meta_transaltion}
\end{figure*}

\begin{table*}[!hbt]
	\centering
	\caption{Average FID scores ($\times 10$) of various $10$-shot testing tasks. $\leftarrow $  represents the reverse translation direction, such as labels$\leftarrow $photos, and $\rightarrow $ represents the forward translation direction, such as labels$\rightarrow $photos. For each translation direction, the best FID scores are in bold.}
	\scalebox{0.85}
	{
		\begin{tabular}{|c|c|c|c|c|c|c|}
			\hline
			& \multicolumn{2}{c|}{CycleGAN \cite{zhu2017unpaired}} & \multicolumn{2}{c|}{StarGAN \cite{choi2017stargan}} & \multicolumn{2}{c|}{Ours} \\ \hline
			&  $\leftarrow $         & $\rightarrow $         & $\leftarrow $         & $\rightarrow $         & $\leftarrow $        & $\rightarrow $       \\ \hline
			labels$\leftrightarrow $photos &  15.10 $\pm$ 1.72             &   28.69 $\pm$ 1.54             &  22.56 $\pm$ 1.55            &    39.27 $\pm$ 1.48           & \textbf{12.16 $\pm$ 1.19}            &   \textbf{26.08 $\pm$ 1.25}          \\ \hline
			horses$\leftrightarrow $zebras & 31.96 $\pm$ 1.63     & 31.45 $\pm$ 2.29       &  34.94 $\pm$ 1.72            &   35.83 $\pm$ 1.81     &      \textbf{31.80 $\pm$ 1.28}       &   \textbf{30.68 $\pm$ 2.09}          \\ \hline
			summer$\leftrightarrow $winter & 23.43 $\pm$ 2.12              &  19.24 $\pm$ 2.02             & 32.18 $\pm$ 1.98             &  34.95 $\pm$ 1.89             &  \textbf{21.50 $\pm$ 1.71}           &  \textbf{17.95 $\pm$ 1.93}           \\ \hline
			apple$\leftrightarrow $orange &  33.16$\pm$ 1.83             &  34.30 $\pm$ 1.51            &    38.93 $\pm$ 1.69         &    43.58 $\pm$ 1.87          &   \textbf{31.34 $\pm$ 1.57}          &   \textbf{31.93 $\pm$ 1.92 }        \\ \hline
			monet$\leftrightarrow $photo & 18.97 $\pm$ 1.75              & 19.93 $\pm$ 2.21              &  33.38 $\pm$ 2.27            &   35.95 $\pm$ 2.18            &   \textbf{18.42 $\pm$ 2.11}          &  \textbf{19.73 $\pm$ 2.08}           \\ \hline
			cezanne$\leftrightarrow $photo &  22.85 $\pm$ 1.71            &   23.24  $\pm$ 1.89          &     29.23  $\pm$ 1.85       & 31.37  $\pm$ 1.97            &   \textbf{22.00 $\pm$ 1.66}         &   \textbf{22.57  $\pm$ 1.78}        \\ \hline
			ukiyoe$\leftrightarrow $photo &  22.90 $\pm$ 2.05            &  21.16   $\pm$ 2.14          &    24.87 $\pm$ 2.25         &   28.99 $\pm$ 2.23           & \textbf{21.88 $\pm$ 2.18 }          &   \textbf{20.86$\pm$ 1.96}          \\ \hline
			vangogh$\leftrightarrow $photo & 22.65 $\pm$ 2.29               &  19.26 $\pm$ 2.18             &  34.91 $\pm$ 2.26            & 36.68 $\pm$ 2.32              &   \textbf{21.57 $\pm$ 2.14}           &   \textbf{18.42 $\pm$ 2.05}          \\ \hline
			photos$\leftrightarrow $maps &   25.57 $\pm$ 1.96           &    27.17 $\pm$ 1.72          &  35.61 $\pm$ 1.88           &  31.36 $\pm$ 1.95            &  \textbf{19.64 $\pm$ 1.64}          &   \textbf{23.44 $\pm$ 1.72 }        \\ \hline
			labels$\leftrightarrow $facades &    23.06 $\pm$ 1.86          &   26.35 $\pm$ 1.73           &  26.55  $\pm$ 2.03          &    32.80  $\pm$ 2.11         &     \textbf{21.72 $\pm$ 2.09}       &   \textbf{25.34 $\pm$ 1.96}         \\ \hline
		\end{tabular}
	}

	\label{table:fid_10shot}
\end{table*}

\section{Experiments}
\subsection{Experimental Setup}
We extensively evaluate the effectiveness and generalizing ability of the proposed MT-GAN algorithm for UDT problem on two kinds of translation task distributions. The first one (denoted as $P_1(T)$) contains $10$ diverse translation tasks collected by \cite{zhu2017unpaired}: labels$\leftrightarrow $photos, horses$\leftrightarrow $zebras, summer$\leftrightarrow $winter, apple$\leftrightarrow $orange, monet$\leftrightarrow $photo, cezanne$\leftrightarrow $photo, ukiyoe$\leftrightarrow $photo, vangogh$\leftrightarrow $photo, photos$\leftrightarrow $maps and labels$\leftrightarrow $facades. In addition, the Facescrub dataset \cite{ng2014data}, which comprises  $531$ different celebrities, is utilized as another collection of domain translation tasks (denoted as $P_2(T)$) that are less diverse, in which different identities are viewed as different domains.  Then we can sample arbitrary two identities to form a  two-domain translation task that aims to transfer the identity of face images while preserving original face orientation and expression.

In our experiments, for both $P_1(T)$ and $P_2(T)$, we simulate the meta domain translation scenarios by randomly select $N=9$ tasks as a training dataset and select the other $1$ task as the testing dataset/task. This procedure could be seen as task-level 10 fold cross-validation. We establish $10$ training datasets and $10$ corresponding testing datasets for both $P_1(T)$ and $P_2(T)$.  For each training dataset, we randomly select overall $2000$ meta batches from the $9$ tasks for model training. We set the meta-batch size $J$ to $2$ to fit the memory limit of the GPU. Following the common settings of few-shot learning, we mainly focus on $5$-shot domain translation and $10$-shot domain translation in all experiments. Moreover, we set the query set's size $L$ to $10$. For the testing task, we randomly select $5$ meta batches from the $1$ task for model testing.

For each meta-training period, we use stochastic gradient descent (SGD) with learning rate $\alpha = 0.0001$ to fine-tune the generators and discriminators. At meta-optimization time, we use the Adam optimizer \cite{kingma2014adam} with learning rate $\beta =0.0002$ to update both meta-generator and meta-discriminator. For model fine-tuning on the testing tasks, we also use the Adam optimizer with learning rate $\beta =0.0002$ to fine-tune both meta-generator and meta-discriminator. The overall iteration number of meta-training $T$ is set to $100$. We set the loss function balance parameters $\lambda_{\text{cyc}}$ and $\lambda_{\text{idt}}$ to be $10$ and $5$.

For each $K$-shot domain translation task, we fine-tune the trained MT-GAN on each meta batch of the testing dataset, and report the average score and its standard deviation. The Frechet Inception Distance (FID) \cite{NIPS2017_7240} that measures similarity between generated image dataset and real image dataset is used to evaluate translation results' quality. The lower the FID is, the better the translation results are. In addition, we perform face classification experiments on  face identity translation tasks. We re-trained $\text{VGG-16}$ network \cite{simonyan2014very} on  Facescrub, and compute the top-1 and top-5 classification accuracy rates of the translation results.

\begin{table*}[!hbt]
	\centering
\caption{Average classification accuracy of $5$-shot and $10$-shot face identity translation tasks. The best top-1 and top-5 classification accuracy are in bold.}
	\begin{tabular}{|c|c|c|c|c|}
		\hline
		&       & CycleGAN \cite{zhu2017unpaired} & StarGAN \cite{choi2017stargan} & Ours \\ \hline
		\multirow{2}{*}{5-shot}  & Top-1 &   11.21 $\pm$ 0.91\%       &  4.36 $\pm$ 0.89\%       &  \textbf{13.05 $\pm$ 1.06\% }   \\ \cline{2-5}
		& Top-5 &  37.12 $\pm$ 1.63\%        &    15.35 $\pm$ 1.11\%    &   \textbf{40.02 $\pm$ 1.22\%}    \\ \hline
		\multirow{2}{*}{10-shot} & Top-1 &  19.04 $\pm$ 1.21\%       &     10.38 $\pm$ 0.78\%    &  \textbf{21.12 $\pm$ 1.03\% }   \\ \cline{2-5}
		& Top-5 &  45.56 $\pm$ 1.56\%     &    37.34 $\pm$ 1.43\%     &   \textbf{48.27 $\pm$ 1.54\%}   \\ \hline
	\end{tabular}
	
	\label{table:cls_5shot_10shot}
\end{table*}

\subsection{Model configuration}
We follow~\cite{zhu2017unpaired} to configure the models. The meta-generator $G$ network consists of two convolution layers with stride $2$ and kernel size $3\times 3$, six residual blocks \cite{he2016deep} with kernel size $3\times 3$ and two transposed convolution layers with stride $0.5$ and kernel size $3\times 3$. For meta-discriminator $D$, we use PatchGANs \cite{isola2016image} that consists of five convolution layers with stride $2$ and kernel size $4\times 4$. For both $G$ and $D$, we use batch normalization \cite{ioffe2015batch} among network layers.

\subsection{Results}
\textbf{Qualitative and Quantitative Evaluation}
We compare our method with two baseline domain translation models, i.e., CycleGAN \cite{zhu2017unpaired} and StarGAN \cite{choi2017stargan}. We retrain both CycleGAN and StarGAN on each meta-batch in testing dataset of a given $K$-shot domain translation task, and report their average performance with the retrained models. We show qualitative comparison results of the testing tasks in Figure \ref{fig:meta_transaltion}. We observe that StarGAN typically produces quite blurry and noisy outputs, and obviously suffers from the limited training samples. CycleGAN maintains the main structure of the source inputs in most cases and transfers some domain-specific features of the target domains in the translation results. However, CycleGAN still fails to locate the accurate regions that domain-specific features should be transferred in, and produces unnatural images. For example, the translated apple by CycleGAN in $10$-shot orange$\rightarrow$apple is surrounded by inaccurate apple features, and the translated map by CycleGAN in $10$-shot photos$\rightarrow$maps mistakenly transfers the land and houses to water label. On the contrary, most of our results well preserve the domain-invariant features \cite{lin2018conditional,huang2018multimodal} and accurately transfer the domain-specific features \cite{lin2018conditional,huang2018multimodal} in the translation results. It should be noticed that, even with limited unpaired training samples, our model is still able to detect semantic regions of source inputs. For instance, our model successfully detects the water area and land in 10-shot domain translation of the Figure \ref{fig:meta_transaltion} (f) and (g).

For the quantitative evaluation, we present the FID score results of various testing tasks in Table \ref{table:fid_10shot}. For face identity translation, we report the top-1 and top-5 face recognition accuracy of generated images from CycleGAN, StarGAN and our model in Table~\ref{table:cls_5shot_10shot}. We observe that the quantitative results are quite related to the qualitative results in Figure \ref{fig:meta_transaltion}, in which our model consistently outperforms CycleGAN and StarGAN. We also find that improvement brought by our model on some natural image generation tasks, such as four painting$\leftrightarrow$photo tasks, is less significant than other tasks, such as photos$\leftrightarrow$maps and labels$\leftrightarrow$photos. Such result is not surprising because only limited samples are hard to include all patterns for natural image generation, while the patterns of photos$\leftrightarrow$maps or labels$\leftrightarrow$photos are more simple and regularized.

Comparing the performance of MT-GAN on $5$-shot and $10$-shot domain translation tasks, we can see that the proposed meta-learning approach is quite robust to the drop in the amount of training samples.  With only $5$ training samples, MT-GAN still successfully transfers source inputs to target domains in most cases. With the increase of training samples, MT-GAN steadily improves performance on $10$-shot domain translation tasks compared with CycleGAN and StarGAN.



\begin{figure*}[t!]
	\centering
	\centerline{\includegraphics[width=15.0cm]{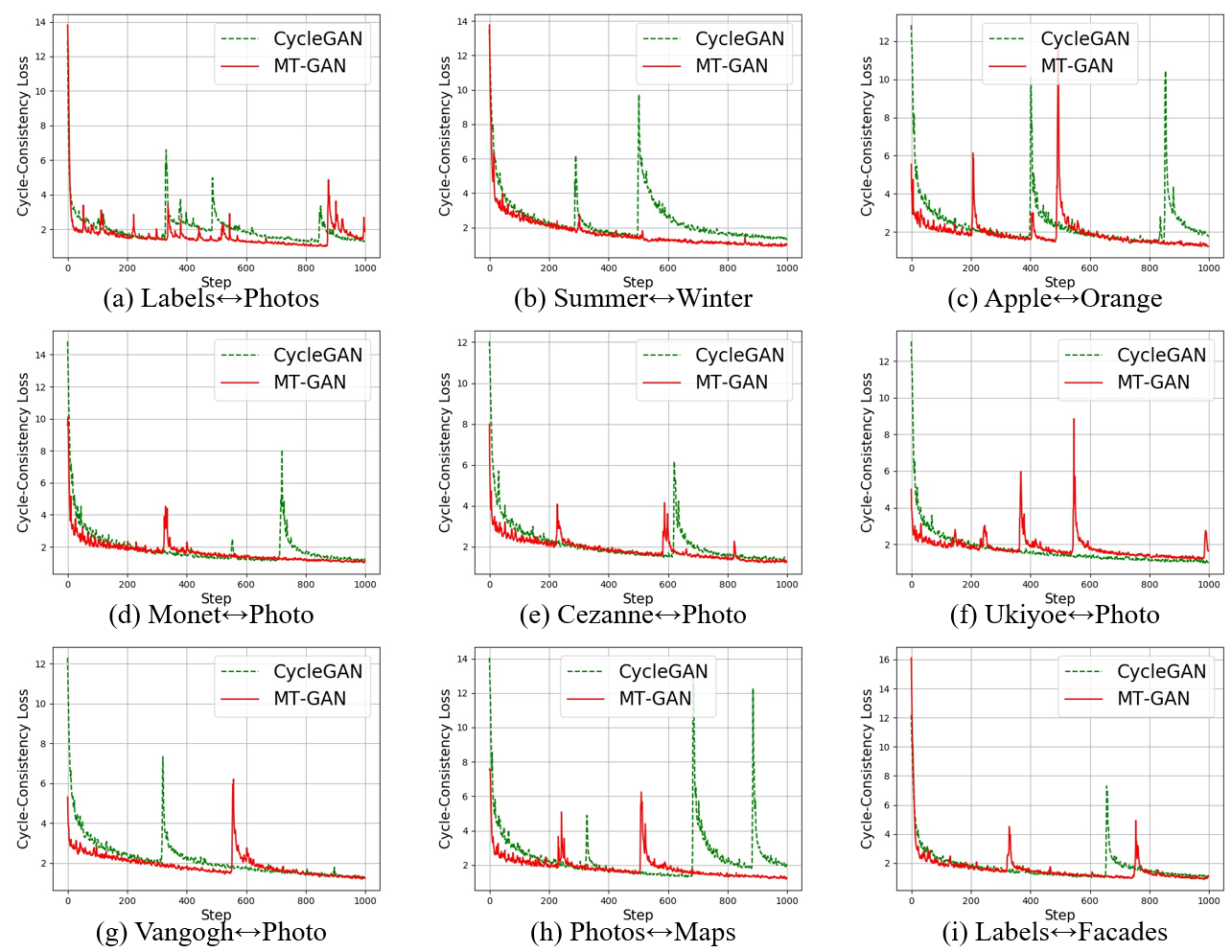}}
	\caption{The training curves of cycle-consistency loss with respect to training step on the different testing tasks.}
	\label{fig:training_curve}
\end{figure*}

\textbf{Convergence Rate}
Our meta-learning based approach has demonstrated that, with several domain translation tasks, it can incorporate the prior learning experiences on these tasks and generalize to a new task with better performance than ordinary translation models in the above experiments. We show that meta-learning brings another benefit, i.e., faster convergence rate in a training process. We show the training curves of cycle-consistency loss with respect to training steps on the different testing tasks in Figure~\ref{fig:training_curve}. We choose cycle-consistency loss to reflect the convergence rate because a smaller cycle-consistency loss indicates that the two translation models well relate two domains. Comparing training curves of CycleGAN and our model, we observe that our model rapidly minimizes the cycle-consistency loss in the first several steps. In addition, our model achieves lower cycle-consistency loss than CycleGAN after numerous iteration steps in most cases. When training abnormality of GAN occurs, we see that our model can recover to original training states more quickly than CycleGAN. These results demonstrate that our model indeed learns adaptation strategies from previous translation tasks, and helps to converge more quickly in current tasks.

\textbf{Saturation Study}
In order to investigate when our model will saturate with the increasing shots, we compare our model with CycleGAN which has shown superior performance in many-shot domain translation. We conduct the experiment on the face identity translation task. As shown in Table \ref{table:more_shots}, we find that our model consistently outperforms CycleGAN in all shots, and performance of two models becomes closer with more shots. This is because with more training samples, a model is able to learn from scratch without prior experiences.

\begin{table}[!h]
\centering
\caption{Average top-1 classification accuracy of face identity translation tasks with different shots.}
\begin{tabular}{|c|c|c|c|c|}
\hline
                &  10-shot & 15-shot & 20-shot & 25-shot \\ \hline
{CycleGAN}   &  19.04\%        & 25.95\%         &  31.84\%   & 35.62\%\\ \hline

{Ours}  &  21.12\%    &  27.63\%     &   32.96\%  & 36.03\%\\ \hline
{Improvement}  &  2.08\%    &  1.68\%     &   1.12\%  &  0.41\% \\ \hline
\end{tabular}

\label{table:more_shots}
\end{table}

\section{Conclusions}
In this work, we manage the unsupervised domain translation (UDT) problem from a meta-learning perspective which aims to effectively incorporate prior domain translation experiences. Accordingly, we propose a model called MT-GAN to find the initialization of a meta-generator and a meta-discriminator that can be used for initialization of any translation task. We jointly train two meta-learners in an adversarial and dual form. We demonstrate our model on ten diverse domain translation tasks and face identity translation tasks. Both qualitative and quantitative results show that the meta-learning based approach significantly outperforms ordinary translation models. In addition, we show that our model can achieve faster convergence rate than CycleGAN, which further demonstrates MT-GAN indeed learns adaptation strategies from previous learning experiences.

For future works, it would be interesting to extend the training paradigm of MT-GAN to other image generation or domain transfer learning tasks. In addition, how to learn the adaptation strategies from many-shot domain translation tasks will be worthy to explore.

{\small
	\bibliographystyle{unsrt}
	\bibliography{Bibliography-File}

\begin{thebibliography}{10}

\bibitem{zhu2017unpaired}
Jun-Yan Zhu, Taesung Park, Phillip Isola, and Alexei~A. Efros.
\newblock Unpaired image-to-image translation using cycle-consistent
  adversarial networks.
\newblock In {\em The IEEE International Conference on Computer Vision (ICCV)},
  Oct 2017.

\bibitem{choi2017stargan}
Yunjey Choi, Minje Choi, Munyoung Kim, Jung-Woo Ha, Sunghun Kim, and Jaegul
  Choo.
\newblock Stargan: Unified generative adversarial networks for multi-domain
  image-to-image translation.
\newblock In {\em The IEEE Conference on Computer Vision and Pattern
  Recognition (CVPR)}, June 2018.

\bibitem{finn2017model}
Chelsea Finn, Pieter Abbeel, and Sergey Levine.
\newblock Model-agnostic meta-learning for fast adaptation of deep networks.
\newblock In {\em Proceedings of the 34th International Conference on Machine
  Learning-Volume 70}, pages 1126--1135. JMLR. org, 2017.

\bibitem{goodfellow2014generative}
Ian Goodfellow, Jean Pouget-Abadie, Mehdi Mirza, Bing Xu, David Warde-Farley,
  Sherjil Ozair, Aaron Courville, and Yoshua Bengio.
\newblock Generative adversarial nets.
\newblock In {\em Advances in neural information processing systems}, pages
  2672--2680, 2014.

\bibitem{he2016dual}
Di~He, Yingce Xia, Tao Qin, Liwei Wang, Nenghai Yu, Tieyan Liu, and Wei-Ying
  Ma.
\newblock Dual learning for machine translation.
\newblock In {\em Advances in Neural Information Processing Systems}, pages
  820--828, 2016.

\bibitem{ng2014data}
Hong-Wei Ng and Stefan Winkler.
\newblock A data-driven approach to cleaning large face datasets.
\newblock In {\em Image Processing (ICIP), 2014 IEEE International Conference
  on}, pages 343--347. IEEE, 2014.

\bibitem{mirza2014conditional}
Mehdi Mirza and Simon Osindero.
\newblock Conditional generative adversarial nets.
\newblock arXiv:1411.1784, 2014.

\bibitem{radford2015unsupervised}
Alec Radford, Luke Metz, and Soumith Chintala.
\newblock Unsupervised representation learning with deep convolutional
  generative adversarial networks.
\newblock arXiv:1511.06434, 2015.

\bibitem{pumarola2018ganimation}
A.~Pumarola, A.~Agudo, A.M. Martinez, A.~Sanfeliu, and F.~Moreno-Noguer.
\newblock Ganimation: Anatomically-aware facial animation from a single image.
\newblock In {\em Proceedings of the European Conference on Computer Vision
  (ECCV)}, 2018.

\bibitem{wang2018high}
Ting-Chun Wang, Ming-Yu Liu, Jun-Yan Zhu, Andrew Tao, Jan Kautz, and Bryan
  Catanzaro.
\newblock High-resolution image synthesis and semantic manipulation with
  conditional gans.
\newblock In {\em Proceedings of the IEEE Conference on Computer Vision and
  Pattern Recognition}, pages 8798--8807, 2018.

\bibitem{wu2016learning}
Jiajun Wu, Chengkai Zhang, Tianfan Xue, Bill Freeman, and Josh Tenenbaum.
\newblock Learning a probabilistic latent space of object shapes via 3d
  generative-adversarial modeling.
\newblock In {\em Advances in neural information processing systems}, pages
  82--90, 2016.

\bibitem{yu2018generative}
Jiahui Yu, Zhe Lin, Jimei Yang, Xiaohui Shen, Xin Lu, and Thomas~S Huang.
\newblock Generative image inpainting with contextual attention.
\newblock In {\em Proceedings of the IEEE Conference on Computer Vision and
  Pattern Recognition}, pages 5505--5514, 2018.

\bibitem{isola2016image}
P.~{Isola}, J.~{Zhu}, T.~{Zhou}, and A.~A. {Efros}.
\newblock Image-to-image translation with conditional adversarial networks.
\newblock In {\em 2017 IEEE Conference on Computer Vision and Pattern
  Recognition (CVPR)}, pages 5967--5976, July 2017.

\bibitem{Yi_2017_ICCV}
Zili Yi, Hao Zhang, Ping Tan, and Minglun Gong.
\newblock Dualgan: Unsupervised dual learning for image-to-image translation.
\newblock In {\em The IEEE International Conference on Computer Vision (ICCV)},
  Oct 2017.

\bibitem{kim2017learning}
Taeksoo Kim, Moonsu Cha, Hyunsoo Kim, Jung~Kwon Lee, and Jiwon Kim.
\newblock Learning to discover cross-domain relations with generative
  adversarial networks.
\newblock In {\em Proceedings of the 34th International Conference on Machine
  Learning}, pages 1857--1865, 2017.

\bibitem{liu2018unified}
Alexander~H. Liu, Yen-Cheng Liu, Yu-Ying Yeh, and Yu-Chiang~Frank Wang.
\newblock A unified feature disentangler for multi-domain image translation and
  manipulation.
\newblock In S.~Bengio, H.~Wallach, H.~Larochelle, K.~Grauman, N.~Cesa-Bianchi,
  and R.~Garnett, editors, {\em Advances in Neural Information Processing
  Systems 31}, pages 2595--2604. Curran Associates, Inc., 2018.

\bibitem{benaim2018one}
Sagie Benaim and Lior Wolf.
\newblock One-shot unsupervised cross domain translation.
\newblock In {\em Advances in Neural Information Processing Systems}, pages
  2104--2114, 2018.

\bibitem{schmidhuber1987evolutionary}
J{\"u}rgen Schmidhuber.
\newblock {\em Evolutionary principles in self-referential learning, or on
  learning how to learn: the meta-meta-... hook}.
\newblock PhD thesis, Technische Universit{\"a}t M{\"u}nchen, 1987.

\bibitem{bengio1990learning}
Yoshua Bengio, Samy Bengio, and Jocelyn Cloutier.
\newblock {\em Learning a synaptic learning rule}.
\newblock Universit{\'e} de Montr{\'e}al, D{\'e}partement d'informatique et de
  recherche~, 1990.

\bibitem{bengio1992optimization}
Samy Bengio, Yoshua Bengio, Jocelyn Cloutier, and Jan Gecsei.
\newblock On the optimization of a synaptic learning rule.
\newblock In {\em Preprints Conf. Optimality in Artificial and Biological
  Neural Networks}, pages 6--8. Univ. of Texas, 1992.

\bibitem{vinyals2016matching}
Oriol Vinyals, Charles Blundell, Timothy Lillicrap, and Daan Wierstra.
\newblock Matching networks for one shot learning.
\newblock In {\em Advances in neural information processing systems}, pages
  3630--3638, 2016.

\bibitem{snell2017prototypical}
Jake Snell, Kevin Swersky, and Richard Zemel.
\newblock Prototypical networks for few-shot learning.
\newblock In {\em Advances in Neural Information Processing Systems}, pages
  4077--4087, 2017.

\bibitem{andrychowicz2016learning}
Marcin Andrychowicz, Misha Denil, Sergio Gomez, Matthew~W Hoffman, David Pfau,
  Tom Schaul, Brendan Shillingford, and Nando De~Freitas.
\newblock Learning to learn by gradient descent by gradient descent.
\newblock In {\em Advances in Neural Information Processing Systems}, pages
  3981--3989, 2016.

\bibitem{mishra2017simple}
Nikhil Mishra, Mostafa Rohaninejad, Xi~Chen, and Pieter Abbeel.
\newblock A simple neural attentive meta-learner.
\newblock arXiv:1707.03141, 2017.

\bibitem{hsu2018unsupervised}
Kyle Hsu, Sergey Levine, and Chelsea Finn.
\newblock Unsupervised learning via meta-learning.
\newblock arXiv:1810.02334, 2018.

\bibitem{metz2018meta}
Luke Metz, Niru Maheswaranathan, Brian Cheung, and Jascha Sohl-Dickstein.
\newblock Meta-learning update rules for unsupervised representation learning.
\newblock arXiv:1804.00222, 2018.

\bibitem{de2019optimal}
Emmanuel de~Bézenac, Ibrahim Ayed, and Patrick Gallinari.
\newblock Optimal unsupervised domain translation.
\newblock arXiv:1906.01292, 2019.

\bibitem{kingma2014adam}
Diederik Kingma and Jimmy Ba.
\newblock Adam: A method for stochastic optimization.
\newblock arXiv:1412.6980, 2014.

\bibitem{NIPS2017_7240}
Martin Heusel, Hubert Ramsauer, Thomas Unterthiner, Bernhard Nessler, and Sepp
  Hochreiter.
\newblock Gans trained by a two time-scale update rule converge to a local nash
  equilibrium.
\newblock In I.~Guyon, U.~V. Luxburg, S.~Bengio, H.~Wallach, R.~Fergus,
  S.~Vishwanathan, and R.~Garnett, editors, {\em Advances in Neural Information
  Processing Systems 30}, pages 6626--6637. Curran Associates, Inc., 2017.

\bibitem{simonyan2014very}
Karen Simonyan and Andrew Zisserman.
\newblock Very deep convolutional networks for large-scale image recognition.
\newblock arXiv:1409.1556, 2014.

\bibitem{he2016deep}
Kaiming He, Xiangyu Zhang, Shaoqing Ren, and Jian Sun.
\newblock Deep residual learning for image recognition.
\newblock In {\em Proceedings of the IEEE conference on computer vision and
  pattern recognition}, pages 770--778, 2016.

\bibitem{ioffe2015batch}
Sergey Ioffe and Christian Szegedy.
\newblock Batch normalization: Accelerating deep network training by reducing
  internal covariate shift.
\newblock arXiv:1502.03167, 2015.

\bibitem{lin2018conditional}
Jianxin Lin, Yingce Xia, Tao Qin, Zhibo Chen, and Tie-Yan Liu.
\newblock Conditional image-to-image translation.
\newblock In {\em Proceedings of the IEEE Conference on Computer Vision and
  Pattern Recognition}, 2018.

\bibitem{huang2018multimodal}
Xun Huang, Ming-Yu Liu, Serge Belongie, and Jan Kautz.
\newblock Multimodal unsupervised image-to-image translation.
\newblock In {\em ECCV}, 2018.

\end{thebibliography}
}
\end{document}